\documentclass[runningheads]{llncs}
\usepackage[T1]{fontenc}
\usepackage{graphicx}
\usepackage{amsmath}
\usepackage{amssymb} 
\usepackage{amsfonts} 
\usepackage{amssymb} 
\usepackage{bm}
\usepackage{mathrsfs}
\usepackage{float} 

\usepackage{pifont}
\usepackage{booktabs}

\usepackage[colorlinks,citecolor=blue,urlcolor=blue, linkcolor=blue,hypertexnames=True]{hyperref}

\begin{document}
%

%
\title{Tackling domain generalization for out-of-distribution endoscopic imaging  }
%
%
\titlerunning{Domain generalization for endoscopic imaging}
 \authorrunning{MA. Teevno et al.}


\author{Mansoor Ali Teevno\inst{1}\orcidID{0000-0002-4669-7267} \and
Gilberto Ochoa-Ruiz\inst{1}\orcidID{0000-0002-9896-8727} \and
Sharib Ali\inst{2}\orcidID{0000-0003-1313-3542}}

\institute{School of Engineering and Sciences, Tecnologico de Monterrey, Mexico\\ \email{gilberto.ochoa@tec.mx} \and
School of Computing, University of Leeds, UK}

%
\maketitle              
\begin{abstract}
 While recent advances in deep learning (DL) for surgical scene segmentation have yielded promising results on single-centre and single-imaging modality data, these methods usually do not generalise to unseen distribution or unseen modalities. Even though human experts can identify visual appearances, DL methods often fail to do so if data samples do not follow the similar data distribution. Current literature for tackling domain gaps in modality changes has been done mostly for natural scene data. However, these methods cannot be directly applied to the endoscopic data as the visual cues are very limited compared to the natural scene data. In this work, we exploit the style and content information in the image by performing instance normalization and feature covariance mapping techniques for preserving robust and generalizable feature representations. Further, to eliminate the risk of removing salient feature representation associated with the objects of interest, we introduce a restitution module within the feature learning ResNet backbone that allows the retention of useful task-relevant features. Our proposed method obtained 13.7\% improvement over the baseline DeepLabv3+ and nearly~8\% improvement on recent state-of-the-art (SOTA) methods for the target (different modality) set of EndoUDA polyp dataset. Similarly, our method achieved 19\% improvement over the baseline and 6\% over best performing SOTA on EndoUDA Barrett’s esophagus (BE) data.    
\end{abstract}

\section{Introduction}
Gastrointestinal (GI) cancer incidence and mortality rate has been on the rise worldwide. In 2020, a total of 1.089 million new cases and 0.769 million deaths were reported due to GI cancer \cite{sung2021global} which makes GI cancer fifth most prevalent type of malignancy and fourth major cause of cancer related mortality. Endoscopy plays a pivotal role in GI cancer screening and diagnosis, however, this tool is quite operator-dependent, and some hard-to-detect cases, causing a considerably high (12\%) missed detection rate on a daily basis \cite{menon2014commonly}. 

In contrast to standard definition (SD), high definition (HD) endoscopes have more advantages such as enhanced field of visualization and higher resolution. In comparison with conventional White light imaging (WLI) modality, narrow-band imaging (NBI) modality utilizes narrow band filters for increased visualization of surface architecture and capillary pattern. Both of these modalities allow clinicians to observe different anatomical information of the same lesion  . For example, NBI modality can be effective in locating hard-to-identify lesions due to the increased visualization surface \cite{pasha2009narrow}. 
\begin{figure}[t]
\center
  \includegraphics[width=3in]{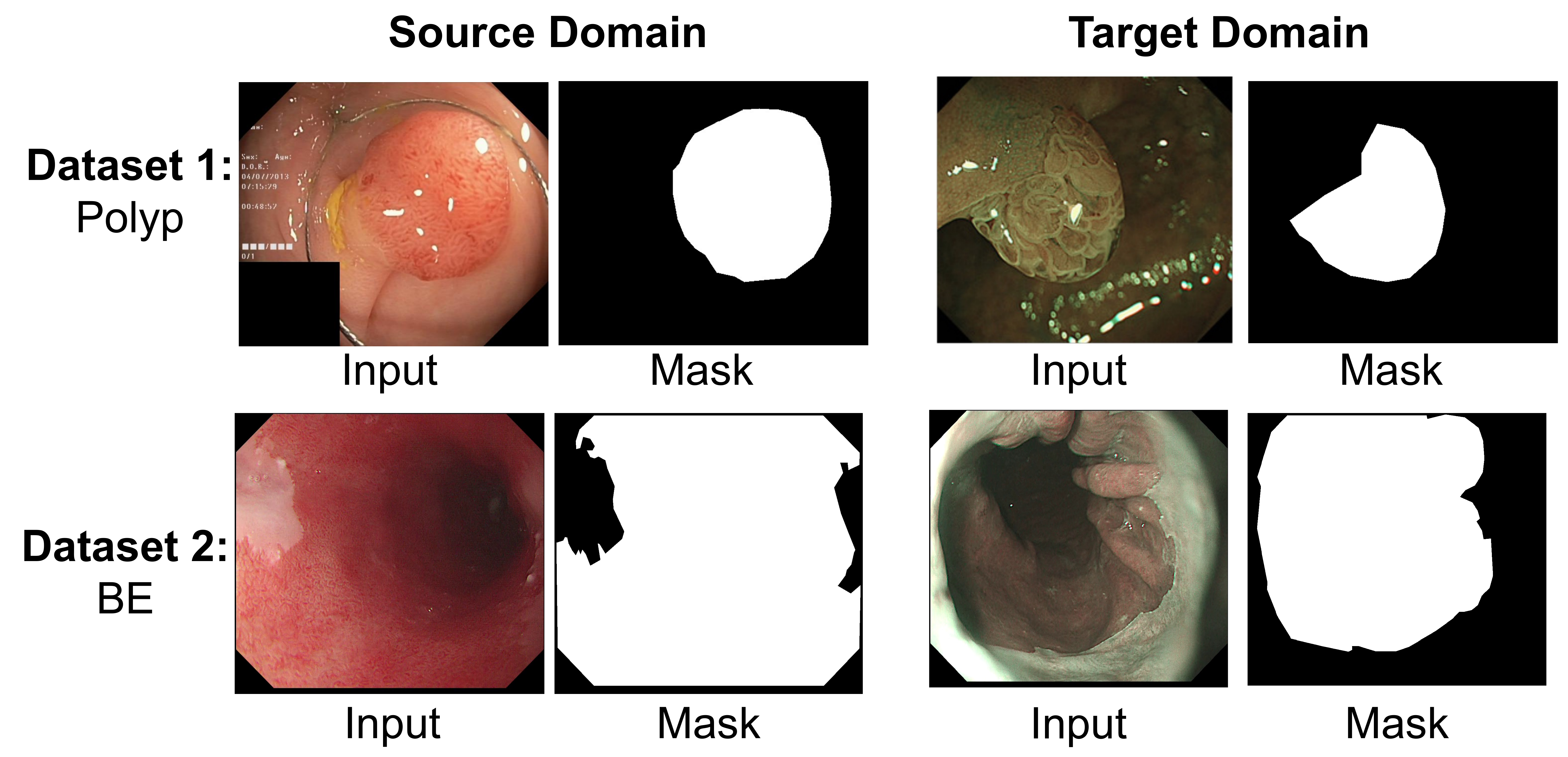}
  \caption{\textbf{Sample images from the EndoUDA dataset.} On the left, these show the images acquired with white light imaging (WLI) and on the right, a narrow-band imaging frames (NBI) for polyps and Barrett's esophagus (BE) \cite{celik_endouda_2021}.}
  \label{fig:endouda}
\end{figure}

Recently, artificial intelligence (AI) driven computer vision methods have been quite successful in computer-aided diagnosis (CADx) systems \cite{bang2021computer}. DL methods have shown promising outcomes in various downstream tasks of endoscopic image analysis such as polyp segmentation \cite{banik2020polyp}, detection \cite{nogueira2022real} and classification \cite{carneiro2020deep} as access to data has become more and more easier and access to vast computing resources. 

Despite the tremendous advancements in developing DL architectures for cancer diagnosis \cite{zhang2019pathologist}, there are still several open challenges which have not been precisely addressed, and specially in the endoscopic domain \cite{jaspers2024robustness}.  A major limitation of the vast majority of current state-of-the-art computer vision models is that they tend to assume that training and test data are sampled from the same data distribution or endoscopic modality. However, data distribution can be very different depending on several factors such as choice of instrument, vendor or data acquisition system to capture the data, lighting conditions, demographics, and the operator. Thus, models trained on one data distribution or modality can perform poorly when encountering out-of-distribution data, as they might not generalize well to the data from distributions different from the training data. Identifying and handling a different test modality samples is critical to design robust and reliable systems for diagnostic procedures that use endoscopic images. Integrating samples from a  different modality in the training set is not feasible as this can skew the learning process due to data imbalance and existing variations thereby making the model sub-optimal to the existing training data modality. Thus, herein we investigate whether or not is possible to design ML models that can perform optimally on samples coming from a different modality than than those of the source domain, with a  minimum performance degradation, i.e., generalizable to unseen modality samples or distribution.

In order to tackle the above-mentioned issues, several works have been carried out in the field of domain generalization (DG) in natural scene \cite{lee2022wildnet,choi2021robustnet} and more recently, some incipient efforts in the endoscopic domain \cite{martinez2023supra,10600804}. To this end, IBN-Net \cite{pan2018two} combined batch and instance norm to learn domain-invariant features, WildNet \cite{lee2022wildnet} proposed to augment source features by adding \textit{wild styles}.
Authors in \cite{kim2022pin} proposed semantic memory to store conceptual knowledge to let the network remember the shared knowledge of each class. In this paper, we propose a DG framework for binary segmentation of polyps and Barrett's esophagus that effectively uses the feature space of the input modality data,  suppresses domain-sensitive information and enhances discriminant features to improve generalizability. The main contribution of this paper is a framework to train models on a given endoscopic modality data that can generalize to other modalities. In order to validate our approach, we tested our methodology on EndoUDA dataset ~\cite{celik2021endouda} which contains two datasets, polyps and Barrett's esophagus containing white light imaging (WLI) and narrow-band imaging (NBI) modalities. Fig. \ref{fig:endouda} shows sample images of EndoUDA dataset. 

A recent study \cite{pan2018two} suggests that DG problem can be addressed by using instance normalization (IN) \cite{ulyanov2016instance}. However, IN alone is not adequate on two counts - 1) It merely standardizes the features while also inevitably removing discriminant features \cite{pan2018two}, and 2) it does not take into account the correlation between channels. Therefore, we propose a novel style normalization and whitening (SRW) block that integrates style normalization and restitution (SNR) block and instance selective whitening (ISW) into the three convolution layers of ResNet-50 backbone of our network to boost DG performance. SNR applies IN after the convolution layers to eliminate style variations and restores the discriminant features. Since, feature covariance contains domain-specific information such as texture and color \cite{gatys2016image,gatys2015texture}, ISW block exploits the feature covariances from the original and augmented image through whitening transformation (WT) \cite{cho2019image} to selectively remove the style information and retain the structure and content. We evaluate the effectiveness of our approach on both same modality test sets and on different modality sets. Specifically, we train the network separately on EndoUDA source (WLI) modalities of polyp and Barrett's esophagus datasets ~\cite{celik2021endouda}. For the evaluation of our method on modality change, we have used target (NBI) modality of EndoUDA polyp and Barrett’s esophagus (BE) datasets~\cite{celik2021endouda}. We will release the implementation code upon acceptance of our work.

The rest of the paper is organized as follows. Section \ref{sec:materials and method} discusses the proposed method, Section \ref{sec:experiments} presents dataset description, implementation details, quantitative and qualitative results and ablation studies. Finally, Section \ref{sec:conclusion} provides the discussion and conclusion. 
\section{Method} \label{sec:materials and method}
\begin{figure}[t]
    \centering
\includegraphics[width=\linewidth]{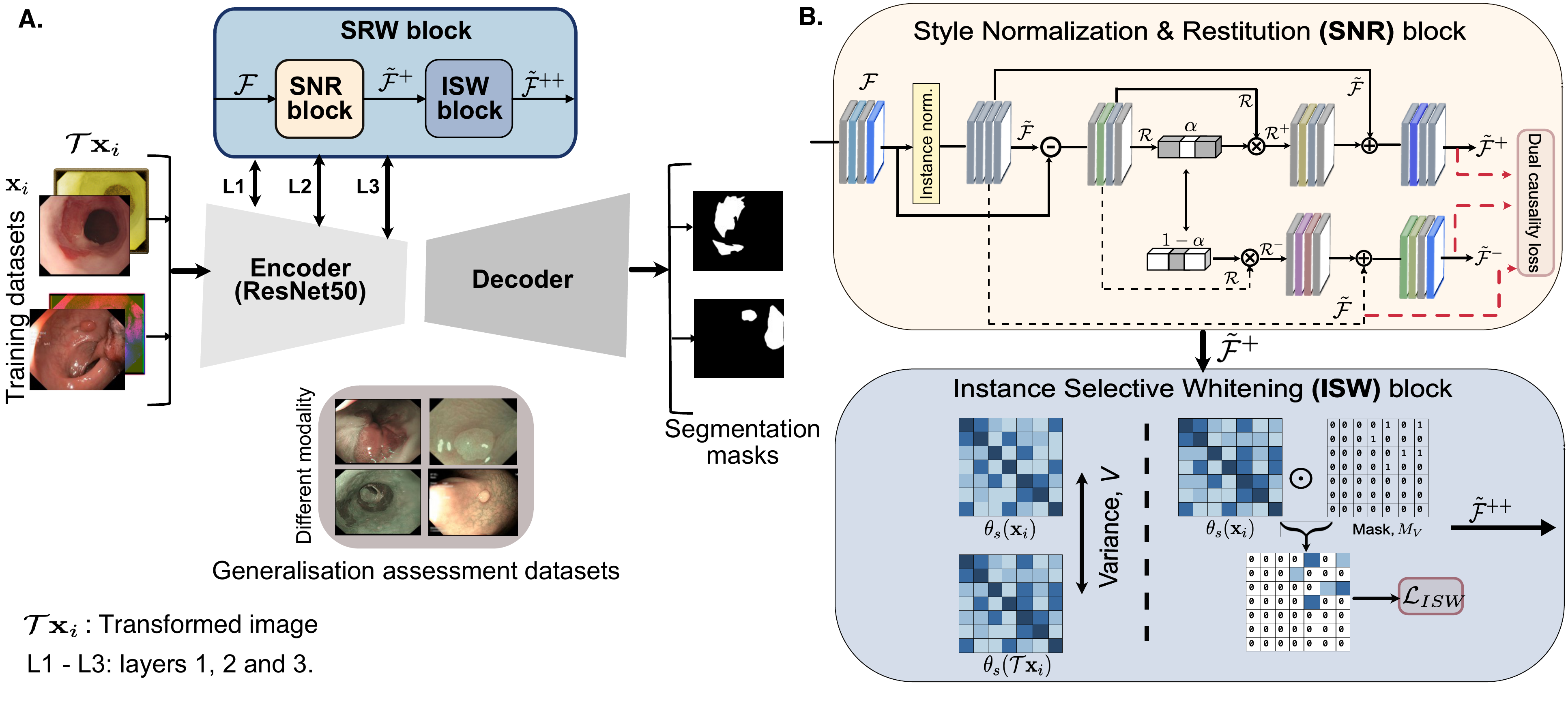}
    \caption{Block diagram of the our proposed method for generalizable surgical scene segmentation. \textbf{A.} depicts the overall flow of the method trained on two datasets. The encoder takes two images, i.e., raw image transformed image, Initially, and feeds the intermediate features to the SRW block.  \textbf{B.} depicts, SNR block \cite{jin2021style} selectively retains the useful features for the generalization, while WT applied selectively suppresses the domain-specific features and and preserves domain-invariant features. Lastly, decoder block performs the up-sampling for segmentation output.  }
    \label{framework}
\end{figure}

The overall proposed framework is described in Fig.~\ref{framework}. First we extract features from ResNet50 backbone which are then passed to the SRW module containing SNR and ISW blocks. In SNR block instance normalization (IN) is applied to standardize the features, while the SNR module recovers the lost domain-invariant information due to IN. Below we detail both SNR and ISW blocks in detail. 
%
%

%
\subsection{Style Normalization and Restitution (SNR block)}
Fig. \ref{framework}.B (top) shows the overall flowchart of the SNR block \cite{jin2021style}. The SNR layer receives an input feature map  $\mathcal{F}\in \mathbb{R}^{N \times C \times H \times W}$ and outputs a enhanced map $\widetilde{\mathcal{F}}^+\in \mathbb{R}^{N \times C \times H \times W}$. Instance normalization is first applied to eliminate style discrepancies in the features maps. It is followed by a restitution step where the task-specific features from the residual are restored (i.e., the difference between style normalized $\widetilde{\mathcal{F}}$ and original feature $\mathcal{F}$ maps). This restoration is carried out by masking  $\mathcal{R}$ with the channel attention vector $\boldsymbol{\mathcal{\alpha}} = [\mathcal{\alpha}_1,\mathcal{\alpha}_2,..\mathcal{\alpha}_C] \in \mathbb{R}^C$. This masking operation is implemented using spatial global average pooling and two fully-connected (FC) layers. Applying channel attention, we obtain 
\begin{equation}
    \mathcal{R}^+(:,:,:,i) = \mathcal{\alpha}_i \mathcal{R}(:,:,:,i)
\end{equation}
\begin{equation}
    \mathcal{R}^-(:,:,:,i) = (1-\mathcal{\alpha}_i) \mathcal{R}(:,:,:,i)
\end{equation}
where $\mathcal{R}(:,:,:,i)$ represents the $i^{th}$ channel of the feature map $\mathcal{R}$.  On the other side of Eqs., (1) and (2), $\mathcal{R}^+$ and $\mathcal{R}^-$ denote the task-relevant and irrelevant features, respectively. Enhanced ($\widetilde{\mathcal{F}}^+$) and corrupted features $(\widetilde{\mathcal{F}}^-$) are then obtained by adding $\mathcal{R}^+$ and $\mathcal{R}^-$ to normalized features $\widetilde{\mathcal{F}}$, respectively. We integrate dual causality loss $\mathcal{L}_{dc}$ function which is basically an entropy minimization loss constraint. With the addition of $\mathcal{R}^+$ to $\widetilde{\mathcal{F}}$,  enhanced feature map $\widetilde{\mathcal{F}}^+$ becomes more discriminative resulting in smaller entropy (pixel randomness or uncertainty), in contrast to the corrupted feature maps which have larger entropy. In $\mathcal{L}_{dc}$, first we compute pixel-wise entropy with the function $\mathcal{E}(.) = -p(.)\log{p}$, where p is denotes softmax probability. The $\mathcal{L}_{dc}$ consists of $\mathcal{L}_{SNR}^+$ and $\mathcal{L}_{SNR}^-$ i.e.,  $\mathcal{L}_{SNR}^+$ + $\mathcal{L}_{SNR}^-$,   
\begin{equation}
    \mathcal{L}_{SNR}^+ = \text{Marginloss}(\frac{1}{w\times h} \sum_{j=1}^{h} \sum_{k=1}^{w} \mathcal{E} (\phi(\widetilde{\mathcal{F}}^+(j,k,:)) - \phi(\widetilde{\mathcal{F}}(j,k,:))))
\end{equation}
\begin{equation}
    \mathcal{L}_{SNR}^- = \text{Marginloss}(\frac{1}{w\times h} \sum_{j=1}^{h} \sum_{k=1}^{w} \mathcal{E} (\phi(\widetilde{\mathcal{F}}(j,k,:)) - \phi(\widetilde{\mathcal{F}}^-(j,k,:))))
\end{equation}
where $\phi $ denotes softmax function and w, h represent height and width of feature vector containing probability values. On the other hand, Marginloss $\ln{(1+\textit{exp}(.))}$ represents a monotonically increasing function whose aim is to serve as smoother optimizer to avoid negative loss values. 

\subsection{Instance Selective Whitening (ISW block)}
It has been shown that whitening transformation (WT) can remove domain-specific information and improve the overall DG performance \cite{cho2019image,pan2019switchable}. For a feature map $\mathcal{F}\in \mathbb{R}^{N \times C \times H \times W}$, WT is a linear transformation which standardizes features by de-correlating the channels. It performs this de-correlation by making the feature covariance matrix ($\theta_s$) close to an identity matrix. 
The conventional way of computing WT through eigen value decomposition is highly computationally expensive. Thus  an alternative implementation proposed in GDWCT \cite{cho2019image} is implemented that computes the deep whitening transformation (DWT):
\begin{equation}
    \mathcal{L}_{\mathrm{DWT}} = \mathbb{E} [\| \theta_\mu - \mathrm{I} \|_1]
\end{equation}
where $\mathbb{E}$ denotes the arithmetic mean. The major limitation of WT is that it can distort the object boundaries \cite{li2017universal} or reduce feature discrimination \cite{pan2019switchable} because $\theta_s$ contains both style and content information. Therefore, in our framework work we introduce the ISW block to selectively remove the style while retain structural information (see Figure \ref{framework}.B, bottom). The ISW block takes the enhanced features $\widetilde{\mathcal{F}}^+$ of the original and transformed images from the SNR block. The network is initially trained for $n$ epochs ($n$ was empirically set to 5) to obtain stable covariance matrices. Afterwards, the convariance $\theta_s$ for both feature maps, as well as their variance V are computed as follows: 
\begin{equation}
    \theta_s = \frac{1}{h \times w} (\widetilde{\mathcal{F}}^+)(\widetilde{\mathcal{F}}^{+})^T
\end{equation}

\begin{equation}
    V = \frac{1}{N} \sum_{i=1}^N \frac{1}{2} ((\theta_s (x_i) - \mu_\theta)^2 + (\theta_s(\mathcal{T}x_i) - \mu_\theta)^2 )
\end{equation} 
\noindent where $\mu_\theta$ denotes the mean of both covariance matrices. It is assumed that high variance values in V indicate the presence of style information which must be suppressed. Therefore, a mechanism disentangle and separate such values  was implemented using k-means clustering. This results in two distinct groups, $G_{high}$ (containing domain style) and $G_{low}$ (containing content information). Based on this clustered V, we compute the mask $M_v$ and consequently $\mathcal{L}_{ISW}$ as follows, 
\begin{equation}
    \mathcal{L}_{ISW} = \mathbb{E} [| \theta_s \textstyle\bigodot \mathcal{M}_v |]
\end{equation}
The overall cost function for our proposed approach is given by, 

\begin{equation}
    \mathcal{L}_{total} = \mathcal{L}_{task}+  \sum_i^L \lambda_1 \mathcal{L}_{ISW}^i + \lambda_2 \mathcal{L}_{dc}^i 
\end{equation}
with $\lambda_1$ and $\lambda_2$ denote the weight of ISW and DC losses, respectively, L represets the number of layers and $\mathcal{L}_{task}$ is cross-entropy loss for semantic segmentation.

\section{Experiments and results} \label{sec:experiments}
\subsection{Datasets}


\noindent \textbf{EndoUDA:}
We used EndoUDA~\cite{celik2021endouda}, a gastrointestinal endoscopy dataset for the segmentation of two precancerous anomalies namely, Barrett’s esophagus and polyps. To evaluate the effectiveness of the generalization capability of our method, we used data from two clinically acquired imaging  modalities, wight light imaging and narrow-band imaging for Barrett's esophagus (BE) and polyps. BE dataset consists of 799 images obtained from 68 unique patients, out of which 515 WLI images are used as source domain data (train set:80\%, validation set: 10\%, test set: 10\%), and  284 NBI modality images as target domain are used for testing. EndoUDA polyp contains 1042 endoscopy images acquired from 20 patients, with 1000 WLI modality images used as source domain data (train set:80\%, validation set: 10\%, test set: 10\%) and 42 NBI images used as target domain data. 

\subsection{Experimental setup} 
For the training of our proposed model, we used DeepLabv3+ \cite{chen2018encoder} as the baseline model for semantic segmentation with ResNet50 backbone. We used SGD optimizer and starting learning rate of $1e^{-2}$ and momentum of 0.9 with polynomial learning rate scheduling. We trained our model for 50 epochs with a batch size of 8.  We also used various augmentations such as color jittering, Gaussian blur, random cropping, random horizontal flipping and random scaling. The model was trained using Pytorch framework with 2 NVIDIA Tesla P100-SXM2-16GB GPUs. 

\noindent\textbf{Evaluataion metrics:} Widely used intersection-over-union (IoU), mean accuracy, precision, and recall have been used for comparing the segmentation performance of our method with state-of-the-art approaches.

\begin{table}[t!]
\caption{Table showing intersection over Union (IoU) scores for the SOTA and our proposed method. All the results are presented on source data only trained models. Standard deviations ($\pm$) are also reported. Highest values of the evaluation metric are represented in bold. }\label{tab1:results}
\begin{tabular}{lcccc|cccc}
\hline
\textbf{EndoUDA (polyp)}&\multicolumn{4}{c}{\textbf{Source WLI modality}}& \multicolumn{4}{c}{\textbf{Target NBI modality}}\\
\hline
Method & IoU & Prec. & Rec. & mAcc.& IoU & Prec. & Rec. & mAcc.\\
\toprule
DeepLabv3+(baseline) \cite{chen2018encoder}   & $\mathbf{81.4}\pm0.163$ & \textbf{90.0} & \textbf{90.3}& \textbf{89.1}& $64.2\pm0.155$ & 74.3 & 76.9 & 78.3 \\
\hline
IBN-Net (CVPR'18) \cite{pan2018two}  & $77.0\pm0.164$ & 85.1 & 89.5 & 85.4 & $68.0\pm0.154$ & 74.2& 70.5 & 75.4  \\
\hline
RobustNet (CVPR'21) \cite{choi2021robustnet}   & $78.1\pm0.162$ &  80.8 & 87.1 & 85.6 & $70.2\pm0.148$ & 77.0 & 80.5 & 76.2 \\
\hline
EndoUDA \cite{celik_endouda_2021}   & -- &  -- & -- & -- & $60.5\pm 0.140$ &  72.2 & 70.4 & -- \\
\hline
 Ours  & $79.1\pm0.166$ &  88.2 & 91.5 & 85.7 & $\mathbf{78.3}\pm0.150$ & \textbf{84.8} & \textbf{85.3}& \textbf{81.3} \\
\hline
\textbf{EndoUDA (BE)}& & \\
\toprule
DeepLabv3+(baseline) \cite{chen2018encoder} & $\mathbf{88.2}\pm0.172$ & \textbf{93.2} & \textbf{91.4} & \textbf{91.8}& $60.7\pm0.147$ & 70.5 & 72.7 &  65.4\\
\hline
IBN-Net (CVPR'18) \cite{pan2018two}   & $76.8\pm0.186$ & 81.1 & 85.6& 80.5 & $68.3\pm0.142$ & 74.9 & 76.3 & 77.8\\
\hline
RobustNet (CVPR'21) \cite{choi2021robustnet} &  $84.2\pm0.167$ & 89.7 & 90.1 & 87.7 & $71.3\pm0.138$ & 82.1 & 83.2 & 77.9\\
\hline
EndoUDA \cite{celik_endouda_2021}   & --- &  --- & --- & --- & $73.3\pm 0.140$ &  83.2 & 78.4 & --- \\
\hline
Ours & $84.3\pm0.142$ & 91.3 & 93.8 & 88.0& $\mathbf{79.7}\pm0.137$ & \textbf{88.5} & \textbf{91.6} & \textbf{84.2} \\

\bottomrule
\end{tabular}
\end{table}

\subsection{Results}\label{sec:results-discussion}
\noindent \textbf{Quantitative results}:
Table~\ref{tab1:results} indicates that all other models outperform the baseline DeepLabv3+, while our proposed method outperforms the SOTA in terms of IoU score on both source and target modality datasets.   



\noindent \textbf{Results on EndoUDA (polyp)}: 
On EndoUDA polyp source modality data, our proposed model got 79.1\% IoU score which is 1\% and 2.1\% higher as compared to RobustNet and IBN-Net, while baseline performed 2.3\% better than our approach in terms of IoU score. On the target domain side, our model with 78.3\% IoU outperformed all the the other methods with a significant margin of 8.1\% (RobustNet), 10.3\% (IBN-Net), 17.8\% (EndoUDA) and 14.1\% (baseline).  

\noindent \textbf{Results on EndoUDA (BE)}: 
 On EndoUDA (BE) source modality, the baseline method performed well as compared to other models, however, on the target domain setting, our method performed much better with IoU score of 79.7\% which is 19\% higher than the baseline, 8.4\%, 11.4\% and 6.4\% better than RobustNet, IBN-Net and EndoUDA. 
 
 
 
\begin{figure}[t!]
    \centering
\includegraphics[width=0.7\linewidth]{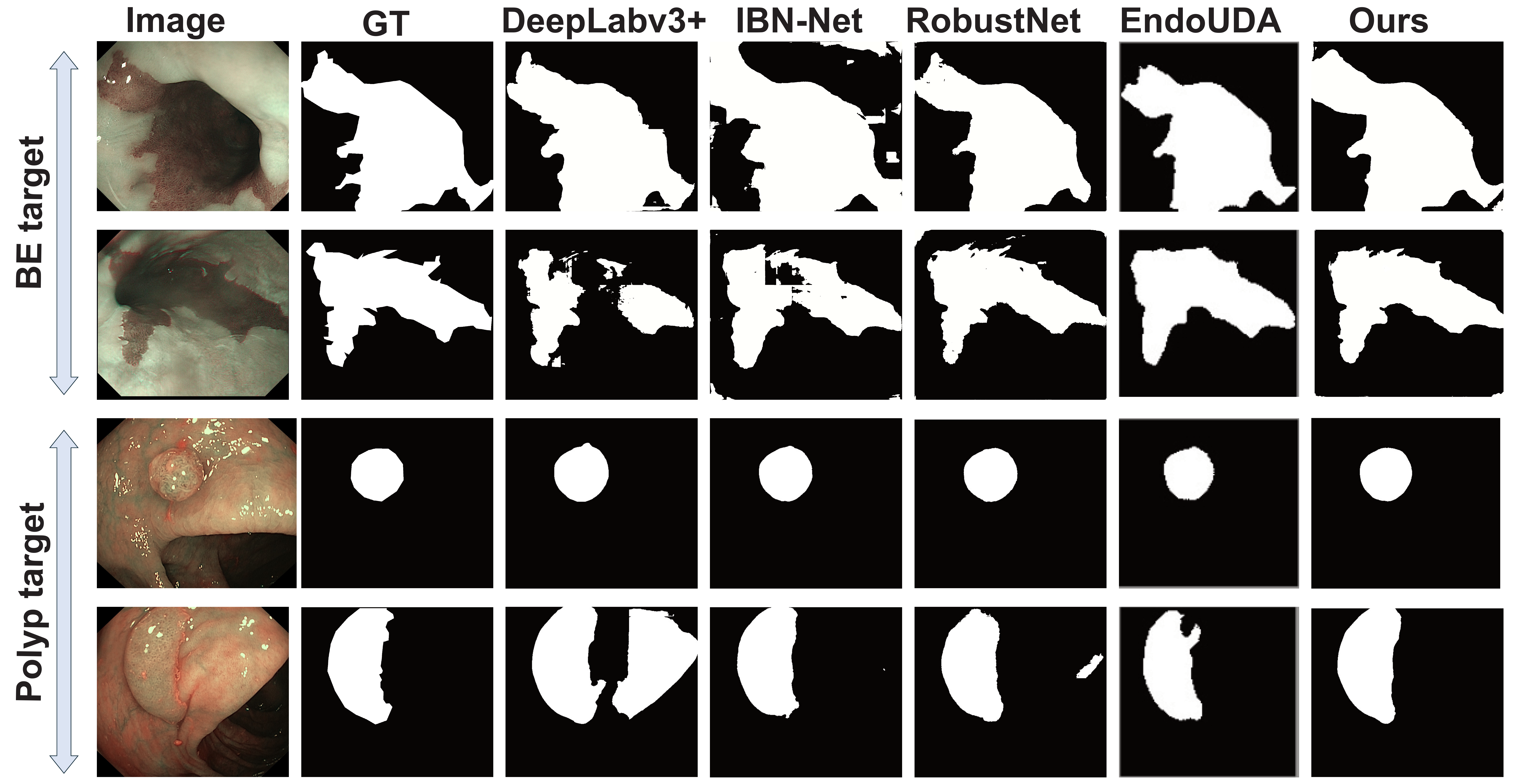}
    \caption{Qualitative results. Top two rows contain qualitative performance on target (NBI) modality Barrett's esophagus data and the bottom two rows consist of results on target (NBI) modality polyp data.}
    \label{fig:qualitative}
\end{figure}

\noindent \textbf{Qualitative results}:
Fig.~\ref{fig:qualitative} presents the qualitative performance on the target NBI modality of both polyp and Barrett's esophagus datasets. For the BE target modality, baseline DeepLabv3+ and the SOTA IBN-Net, RobustNet and EndoUDA struggle to produce clear segmentation boundaries while our proposed method performs better and is closer to the ground truth segmentation mask. Similarly, on the target polyp dataset modality, we can observe on the fourth row that our method does exceptionally well on segmenting the lesions effectively in comparison with all other methods. 

%
%

\subsection{Ablation study} 
Table 3 (\textbf{Supplementary material}) ablates different network configurations where we restrict ISW block to first three layers since deeper layers are not rich in style information. Furthermore, we ablate SNR block between the backbone layers and see how the network behaves.  
Table 4 (\textbf{Supplementary material}) demonstrates behaviour of different number of SNR blocks with fixed ISW blocks to see the effectiveness of our approach. It can be observed that the proposed SRW block (ISW and SNR) configured between layer 1 to layer 3 of ResNet50 backbone produces the best segmentation performance.

\section{Discussion and Conclusion} \label{sec:conclusion}

From Table \ref{tab1:results}, it can be observed that our proposed framework for DG achieves competitive performance on same modality dataset with only the baseline model surpassing in performance, while our model outperformed all the SOTA methods in all the evaluation metric on the target modality showing effectiveness in the generalization performance. On the qualitative side, our method performs well on segmenting lesions with improved segmentation boundaries as compared to other SOTA methods. However, it fails in certain areas such as top-left part of the first row in Fig. \ref{fig:qualitative}. Ablation studies on using SNR and ISW blocks at different ResNet stages indicate that the discriminant information can be retained in the shallow layers (1 to 3) of the network. 

In this work, we addressed the DG problem for surgical scene segmentation by addressing the weaknesses in the IN and WT. We integrated a restitution network to restore the discriminant information lost due to IN, and ISW block to selectively suppress the style information and retain content useful for the model generalizability. We used DeepLabv3+ as the base model for binary segmentation with ResNet50 as the backbone. Results on two different modality datasets showed an improvement in segmentation performance.

\subsubsection{Acknowledgements} The authors wish to acknowledge the Mexican Council for Science and Technology (CONAHCYT) for their support in terms of postgraduate scholarships in this project. This work has been supported by Azure Sponsorship credits granted by Microsoft's AI for Good Research Lab through the AI for Health program. The project was also partially supported by the French-Mexican ANUIES CONAHCYT Ecos Nord grant 322537 and by the Worldwide Universities Network (WUN) funded Project ``Optimise: Novel robust computer vision methods and synthetic datasets for minimally invasive surgery."

%
%
%
\bibliographystyle{splncs04}
\bibliography{mybib}
%




\end{document}